\documentclass[final,3p,times]{elsarticle}
\usepackage{amssymb}
\usepackage{array}
\usepackage{amsmath}
\usepackage{subcaption}
\usepackage{multirow}
\usepackage{pgfplots}
\pgfplotsset{compat=newest}
\usepackage{booktabs}
\usepackage{comment}

\usepackage{xcolor}
\usepackage{color}
\usepackage{colortbl}
\definecolor{Gray}{gray}{0.9}
\definecolor{LGray}{gray}{0.8}
\usepackage{soul}
\usepackage{makecell}
\usepackage{caption}
\usepackage{algorithm}
\usepackage{algpseudocode}
\usepackage{threeparttable}
\usepackage{enumitem}
\usepackage{wasysym}
\usepackage{hyperref}

\usepackage{ulem}

\journal{Neural Networks}
\usepackage[section]{placeins}

\begin{document}

	\begin{frontmatter}
		
		\title{Task-Augmented Cross-View Imputation Network for Partial Multi-View Incomplete Multi-Label Classification}
		\author[address1]{Lian Zhao}
	    \author[address2]{Jie Wen}	    
	    \cortext[cor1]{Corresponding author}
	    \author[address1]{Xiaohuan Lu\corref{cor1}}
	    \ead{xhlu3@gzu.edu.cn}
	    \author[address3,address4]{Wai Keung Wong}

            \author[address1]{Jiang Long}
	    \author[address1]{Wulin Xie}
		\affiliation[address1]{organization={College of Big Data and Information Engineering, Guizhou University},
			city={Guiyang},
			country={China}
		}
			
		\affiliation[address2]{organization={Shenzhen Key Laboratory of Visual Object Detection and Recognition, Harbin Institute of Technology},
			city={Shenzhen},
			country={China}}

             \affiliation[address3]{organization={School of Fashion and Textiles, The Hong Kong Polytechnic University},
			city={Hong Kong},		
			country={China}}
            
            \affiliation[address4]{organization={Hong Kong Laboratory for Artificial Intelligence in Design},
			city={Hong Kong},		
			country={China}}
	
	\begin{abstract}
		In real-world scenarios, multi-view multi-label learning often encounters the challenge of incomplete training data due to limitations in data collection and unreliable annotation processes. The absence of multi-view features impairs the comprehensive understanding of samples, omitting crucial details essential for classification. To address this issue, we present a task-augmented cross-view imputation network (TACVI-Net) for the purpose of handling partial multi-view incomplete multi-label classification. Specifically, we employ a two-stage network to derive highly task-relevant features to recover the missing views. In the first stage, we leverage the information bottleneck theory to obtain a discriminative representation of each view by extracting task-relevant information through a view-specific encoder-classifier architecture. In the second stage, an autoencoder based multi-view reconstruction network is utilized to extract high-level semantic representation of the augmented features and recover the missing data, thereby aiding the final classification task. Extensive experiments on five datasets demonstrate that our TACVI-Net outperforms other state-of-the-art methods. 
	\end{abstract}
	
\end{frontmatter}
	
	\section{Introduction}
In recent years, with the significant progress of representation learning methods, traditional single-view data analysis methods are gradually showing their limitations, and they are difficult to fully adapt to the growing demands of diverse application scenarios \cite{dou2025learning}\cite{jiang2024deep}\cite{lin2023inferring}. Therefore, multi-view data integration has become an important way to break through the inherent barriers of single data sources and alleviate the lack of information \cite{li2020bipartite}\cite{lohith2023multimodal}. This trend is accompanied by the rapid development of data acquisition technology, which makes a huge amount of multi-view data from all kinds of media or with different styles emerge, and providing more possibilities for achieving a comprehensive and accurate description of observation targets \cite{zhao2022multi}\cite{ravi2023systematic}. In short, multi-view data refers to a series of complementary information collections collected from different observation angles for the same object \cite{li2021incomplete}\cite{sun2021wmlrr}. For example, three unique visual information obtained from X-rays, CT scans, and magnetic resonance imaging (MRI) in medical diagnostics constitute a multi-view dataset; retinal images acquired at four different positions can also constitute another multi-view retinal dataset \cite{luo2024deep}. More commonly, multi-view or multi-modal datasets that incorporate multiple types of data including text, images, and videos have been utilized in many application scenarios such as web page classification \cite{lu2019online}\cite{liu2023dicnet}. Concurrently, it is observed that single-label data frequently fail to adequately reflect the intricacies of real-world scenarios \cite{ou2024view}. For instance, a picture of a flower may also require labels such as ``winter'' or ``greenhouse'', demonstrating that multi-label provide a more comprehensive category feature space and more accurately preserve inter-label relationships \cite{chen2023meta}. 
Given the diversity of objects in multi-label classification tasks, researchers have innovatively combined multi-view learning with multi-label classification. This integration not only allows for the full utilization of data from various perspectives, but also provides the chance to capture the complex relationships within target objects and among labels \cite{ma2022joint}. These advantages collectively improve classification performance, making multi-view multi-label classification (MVMLC) a domain of significant research interest \cite{tan2018incomplete}.

For MVMLC, many methods have been proposed in the
past few years, such as the latent semantic aware MvMLC \cite{zhang2018latent}, multi-view probabilistic model LCBM \cite{sun2020lcbm}, and maximizing label-feature dependency based method \cite{zhao2022learning}. However, MVMLC assumes all views and their corresponding labels are complete, a premise that does not always correspond to real-world situations \cite{zhao2024partial}\cite{ren2024novel}. For instance, in the context of video material, audio tracks are often absent, and text descriptions may be incomplete. Similarly, multimedia content disseminated on social media platforms frequently exhibits incomplete annotations owing to varying levels of user engagement. These practical situations give rise to the common challenge of incomplete multi-view features and incomplete category information in real-world datasets, which we define in our paper as ``partial multi-view incomplete multi-label classification (PMVIMLC)'' \cite{pu2024adaptive}. To solve such problem, various approaches for PMVIMLC have been proposed recent years. Tan et al. initially introduced the iMvWL model to construct a shared low-dimensional subspace based on matrix decomposition \cite{tan2018incomplete}. And Li et al. proposed the NAIM3L model to preserve the global consistency and local properties of the label structure \cite{li2021concise}. Nevertheless, since these models employ shallow machine learning techniques to explore the information contained in features and labels, they are constrained in their ability to extract higher-order discriminative features from data. Therefore, Liu et al. developed the DICNet model for extracting deeper semantic representations from multi-view data \cite{liu2023dicnet}. However, the aforementioned models introduce a missing view indication matrix to circumvent unknown views and rely solely on the available data for classification, which fails to fully utilize the potential cross-view complementary information and the intrinsic relationships of the missing data, rendering it suboptimal \cite{liu2023information}. To address this limitation, we propose to construct models that are capable of directly recovering missing data, despite the inherent challenges associated with this task.

In light of such an issue, we propose a task-augmented cross-view imputation network (TACVI-Net) for partial multi-view incomplete multi-label classification. TACVI-Net has been designed to handle arbitrary view-missing scenarios and enhance representation learning through effective view completion. It is acknowledged that raw data may contain task-irrelevant redundant information \cite{xu2023untie}, which may result in the accumulation of such information if used directly to fill in missing data. Therefore, we design a view-specific encoder-classifier architecture based on variational coding, inspired by the information bottleneck theory. This architecture can obtain enhanced features that are highly relevant to the task so that effectively avoiding the impact of irrelevant redundant information on model performance. In the absence of view features, it is of the utmost importance to determine an appropriate filling strategy to provide sufficient and useful information. Given the existence of the shared semantic information across multi-view, we convert each input view into a hidden representation via the view encoder and merge them into a joint representation to form a fused shared representation. Subsequently, the fused shared representation is employed to reconstruct missing data via the view decoder, which is then integrated with the existing data to construct a complete dataset. In this way, we can obtain more sufficient information to exploit the consistency and complementarity across different views, thus improving the final classification performance.

In conclusion, our model TACVI-Net makes the following contributions:

\begin{itemize}
        \item[$\bullet$]We propose a novel two-stage framework that integrates the acquisition of task-relevant 
    information with the completion of missing views. The first stage extracts task-relevant features, while the second stage aims to obtain complete multi-view data to improve the accuracy of the classification process. 
	\item[$\bullet$]Different from previous missing-view recovery approaches, we employ the information bottleneck theory to eliminate task-irrelevant redundant information prior to recovery, effectively preventing the accumulation of noise caused by such information. This guarantees that the model can focus on exploiting and utilizing useful information within the data, thus enhancing the reliability of missing-view recovery.
	\item[$\bullet$]we apply an imputation strategy in autoencoder-based architecture which can increase the amount of available information, thereby strengthening the model's ability to better understand and capture consistent as well as complementary information across multiple views. The experimental results on several datasets demonstrate the superiority of our method.
\end{itemize}	
	
\section{Preliminaries}
We devote this section to the problem definition of PMVIMLC and the rules for representing the associated symbols. Additionally, because our task is closely related to multi-view learning and multi-label classification, we sequentially introduce related works in multi-view multi-label classification and partial multi-view incomplete multi-label classification.

\begin{table*}
	\centering
	\caption{Main notations utilized in our paper.}
	\label{datasets}
	\label{table_one}
	\begin{tabular}{c @{\hspace{70pt}} c}
		\toprule
		Notation & Explanation \\
		\midrule  
		$n$ & the number of all samples  \\
		$m$ & the number of all views  \\
		$c$ & the number of all classes \\
		$d_{l_{x}}$ & the dimension of $X_{l}$  \\
		$d_{l_{v}}$ & the dimension of $V_{l}$, $V_{l}^{'}$ and $T_{l}$\\
		$d_{e}$ & the dimension of $Z_{l}$, $Z_{T}^{l}$, $\widehat{Z}$ and $\widehat{Z_{T}}$  \\
		$X_{l}$ & original data of view $l$ $(R^{n\times d_{l_{x}}})$ \\
		$V_{l}$ & related-task data of view $l$ $(R^{n\times d_{l_{v}}})$ \\
		$V_{l}^{'}$ & reconstructed related-task data of view $l$ $(R^{n\times d_{l_{v}}})$ \\
		$T_{l}$ & complete data of view $l$ $(R^{n\times d_{l_{v}}})$ \\
		$Z_{l}$ & embedding feature of $V_{l}$ $(R^{n\times d_{e}})$  \\
        $\widehat{Z}$ & fusion feature of all imcomplete views $(R^{n\times d_{e}})$\\
		$Z_{T}^{l}$ & embedding feature of $T_{l}$ $(R^{n\times d_{e}})$ \\
		$\widehat{Z_{T}}$ & fusion feature of all complete views $(R^{n\times d_{e}})$ \\
		$Y$ & real label matrix $(\left \{ 0,1 \right \} ^{n\times c} )$ \\
		$Y^{'} $ & predictive label matrix $(\left [ 0,1\right ] ^{n\times c} )$ \\
		$U$ & missing-view indication matrix $(\left \{ 0,1 \right \} ^{n\times m} )$\\
		$G$ & missing-label indication matrix $(\left \{ 0,1 \right \} ^{n\times c} )$\\
		$\backepsilon_{l}$ & encoder for the first stage of $X_{l}$\\
		$\complement _{l}$ & classifier for the first stage of $V_{l}$\\
		$\Psi_{l}$ & encoder for the second stage of $V_{l}$ and $T_{l}$\\
		$\Phi_{l}$ & decoder for obtaining reconstructed feature $V_{l}^{'}$\\
		$\mathbb{C}$ & classifier for the second stage of fusion feature $\widehat{Z_{T}}$\\
		\bottomrule  
	\end{tabular}
\end{table*}

\subsection{Problem Statement and Associated Symbols}
For the PMVIMLC task, we delineate the following notations: $\left \{ X_{l} \right \} _{l=1}^{m}$ and $Y$ symbolize the original multi-view data and corresponding labels, respectively, with $m$ denoting the total number of views. Each $X_{l}= \left \{ x_{l_{1}},x_{l_{2}},...,x_{l_{n}}\right \} \in R^{n\times d_{l_{x}}}$ encapsulates the properties of instances within the $l$-th view, where $n$ represents the instance count and $d_{l_{x}}$ specifies the feature dimensionality for this view. Specifically, $x_{l_{i} }$ denotes the attributes of the $i$-th sample observed through the $l$-th perspective. The label matrix $Y=\left \{ y_{1},y_{2},...,y_{n} \right \} \in\left\{ 0,1 \right \} ^{n\times c}$, with $c$ represents the label count, and $Y_{i, j} = 1$ implies the $i$-th instance carries the $j$-th tag; conversely, $Y_{i, j} = 0$. Specifically, $ y_i $ represents the status of all possible labels for the $ i $-th sample. To further account for missing data, we introduce a missing-view indication matrix $U\in\left \{ 0,1 \right \} ^{n\times m} $, where $U_{i,j} =1$ affirms the presence of the $i$-th sample in the $j$-th view, otherwise $U_{i,j} =0$. Similarly, $G\in\left \{ 0,1 \right \} ^{n\times c} $ serves as the missing-label indication matrix, with $G_{i,j} =1$ signifying awareness of the $j$-th label for the $i$-th instance, or $G_{i,j} =0$ for unknown status. In preprocessing, missing values in feature matrices $\left \{ X_{l} \right \} _{l=1}^{m}$ and label matrix $Y$ are set as 0. Given the above complete or incomplete multi-view training data with complete or incomplete labels, the objective of our PMVIMLC task is to train a model that can accurately predict complete label sets for each multi-view sample with complete or incomplete views. To facilitate readers' rapid identification of essential notations, we present a comprehensive list of the primary notations used in this paper in Table \ref{table_one}.

\subsection{Related Works}
\label{relatedworks}
\subsubsection{Multi-view multi-label classification}
The field of MVMLC focuses on coping with data that contain multiple views and multiple labels at the same time which significantly increasing the complexity. Incorporating multi-view learning strategies has been shown to be more effective in solving such problems compared to multi-label classification approaches that rely only on a single view, inspiring a large increase of related research. For instance, Zhang et al. proposed a method that aligns the latent patterns of different views in the kernel space using matrix factorization, aiming to exploit the complementary information among multiple views \cite{zhang2018latent}.
Wu et al. introduced a method called SIMM, which employs orthogonal constraints on the shared subspace to extract unique information from multiple views, simultaneously optimizing the adversarial loss and the multi-label loss to obtain shared information across views \cite{wu2019multi}. Zhang et al. introduced RAIN, which facilitates interaction and semantic label embedding under the mechanism of multiple heads of attention to enhance label associations and dynamically acquire specific label representations based on learned label embedding \cite{zhang2021relation}. Sun et al. put forward the generative model LCBM to fuse multi-view features and establish inter-label dependencies by using a Bernoulli mixture model \cite{sun2020lcbm}. 

The CDMM method proposed by Zhao et al. employs a nonlinear kernel neural network to create separate classifiers for each view, ensuring inter-view consistency, and incorporates the Hilbert-Schmidt criterion to enhance inter-view diversity \cite{zhao2021consistency}. Yu et al. presented the M3AL method, which combines multi-view self-representation learning to optimize information separation, and employs a unique query strategy to select highly informative sample-label pairs for active learning based on shared, individual information and view differences \cite{yu2021multiview}. Zhao et al. introduced the LVSL method for non-aligned multi-view multi-label classification, aiming to capture view-specific label information and low-level label structure under a unified system. This method exploits structural consistency between views and labels and enhances label spatial information \cite{zhao2022non}. 
The D-VSM approach proposed by Lyu et al. employs DeepGCN to directly encode feature representations at the individual view level, preserving the unique information of each view while facilitating effective coupling of information across views \cite{lyu2022beyond}.
Yin et al. put forward the MMDOM method, which ensures that the low-dimensional shared subspace maintains both local and global data structures through first-order and second-order similarity matrices, effectively protecting manifold structures \cite{yin2023multi}. Xiao et al. developed an SVM-based approach, MVMLP, to obtain consensus information by reinforcing output similarity between views while introducing complementarity by using different views as privileged information for each other \cite{xiao2024new}.

\subsubsection{Partial multi-view incomplete multi-label classification}
Research on the MVMLC task primarily relies on the assumption of data completeness, which does not fully account for the potential for incompleteness in practical applications. As a result, the PMVIMLC task has attracted considerable attention from researchers who have endeavored to develop methods that address the incompleteness problem in multi-view multi-label data. For example, Tan et al. introduced the iMVWL model, which integrates local label association learning, shared subspace and predictor, optimizes cross-view relationships and exploits weak label information \cite{tan2018incomplete}. Zhu et al. developed the GLMVML-IVL model with the objective of learning specific label features, pseudo-class label matrices, low-rank hypothesis matrices, global and local label correlations, and complementary information, thereby providing a more nuanced perspective \cite{zhu2020global}. Li et al. proposed NAIM3L, which further addressed the learning issue of unaligned multi-view multi-label data by aligning all multi-view labels within a common label space \cite{li2021concise}. Qu et al. introduced an active learning method, iMVMAL, which adapts undercomplete autoencoders and uses indicator matrices to handle missing data, effectively learning both shared and individual features of samples \cite{qu2021incomplete}. Liu et al. proposed the IMVPML framework, which employs low-rank and sparse decomposition for label cleaning, constructs a shared subspace across views, and utilizes graph regularization to ensure accurate labeling and orthogonality across subspaces \cite{liu2022incomplete}.

Liu et al. proposed DICNet, which incorporates contrastive learning into the multi-view learning framework. This approach reduces the distance between positive sample pairs while increasing the distance between negative sample pairs, thereby enhancing the model's capacity for information extraction across diverse views \cite{liu2023dicnet}. Liu et al. also proposed the MTD model, which decouples the common single-channel view-level representation in multi-view deep learning methods into shared and view-specific representations, with the aim of fully exploiting the complementary and consistent information across different views \cite{liu2024masked}. Tan et al. advocated for the application of information theory as a means to enhance and refine datasets that are inherently incomplete. The theoretical framework adopted by the researchers allows them to balance the available information to emphasize what is crucial for the task \cite{tan2024two}. The SIP model, proposed by Liu and Xu et al., attempts to compress cross-view representations to maximize the amount of shared information, thereby better predicting semantic labels \cite{liupartial}. 
However, despite the diverse strategies offered by the aforementioned methods for addressing PMVIMLC, three key limitations persist: 1) High computational costs in training certain models. 2) Insufficient modeling of view-label interactions in some approaches, failing to capture complex dependencies. 3) Avoiding or eliminating missing view data, potentially leading to suboptimal information utilization. To overcome these issues, we propose to recover the information related to missing data by jointly exploring the interactions between multiple views and labels, and then using the recovered information as incremental information to break the performance limitations of existing methods.

\begin{figure*}[!t]
	\centering
	\includegraphics[width=5.7in,height=3in]{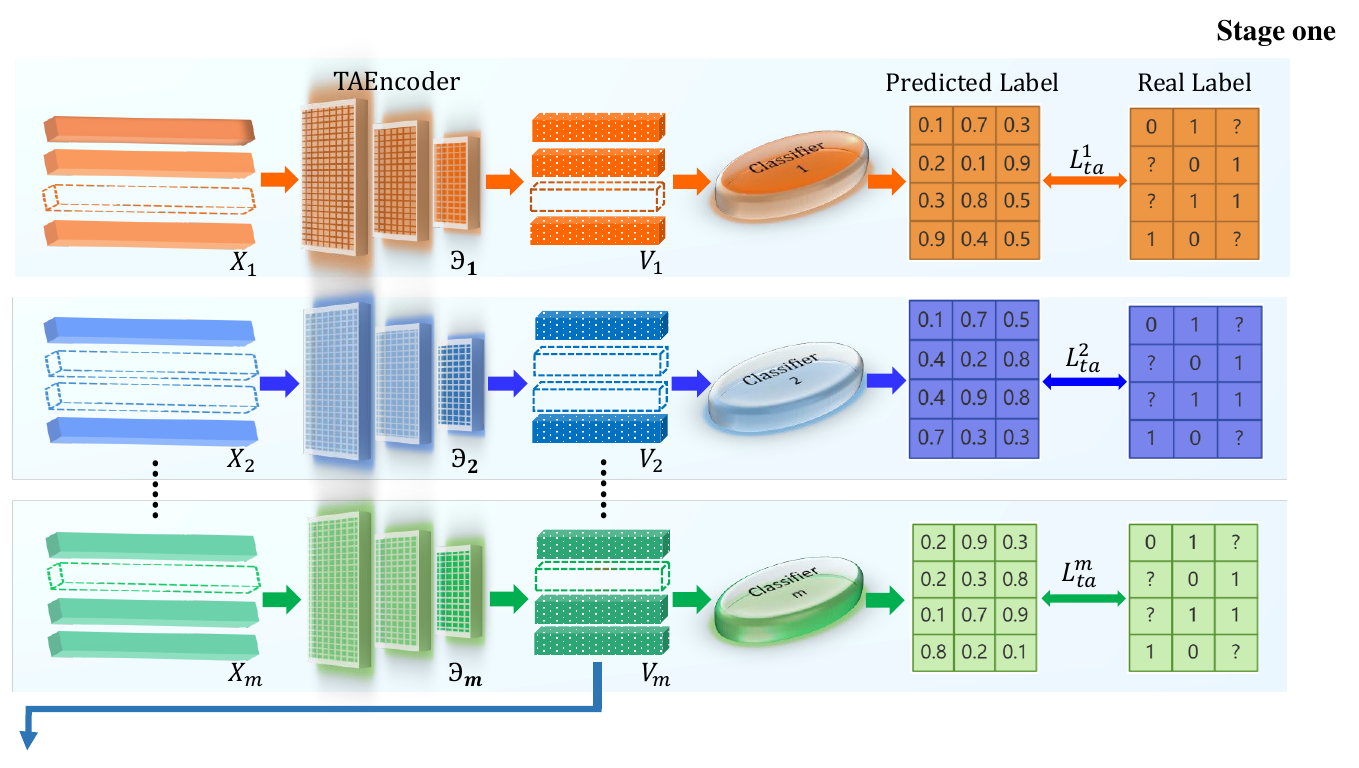}
	\includegraphics[width=6.31in,height=2.6in]{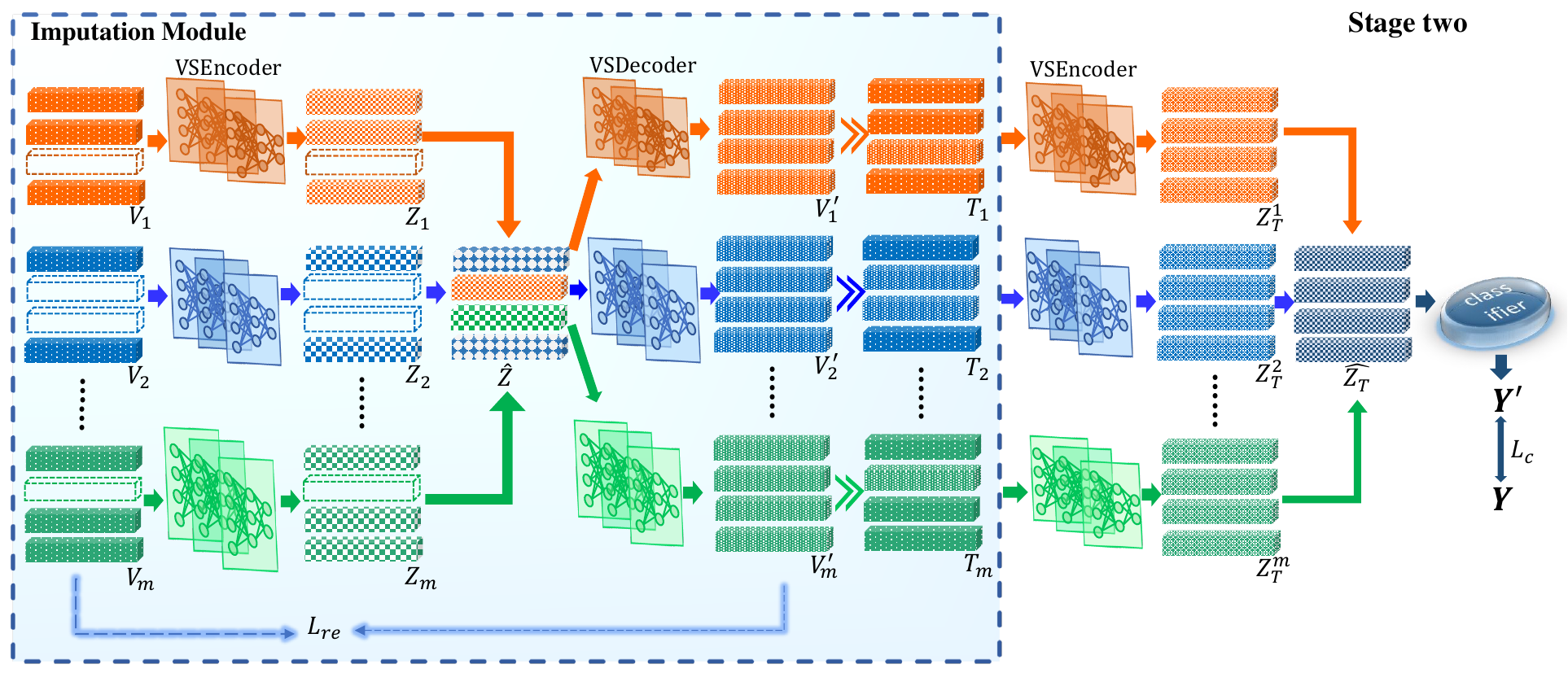}
	\caption{The sketch of our TACVI-Net. The TAEncoder refers to a task-augmented encoder based on variational encoding, while the VSEncoder and VSDecoder denote view-specific encoders and decoders that are designed with a symmetrical structure relative to each other, both of which are based on MLPs.}
	\label{figure_one}
\end{figure*}

\section{The proposed method}
This section presents an in-depth examination of the underlying motivations for the design of our model, accompanied by a comprehensive analysis of the macro-architecture and core components of our approach.
\subsection{Motivation}
For one thing, it is well known that different views share the same high-level semantic information in the classification task because they are different descriptions of the same abstract target. This means that it becomes feasible to infer the missing information backwards based on a learned pattern if we can capture the shared high-level semantic information \cite{liu2023information}. In light of above analysis, we design a cross-view imputation module with an autoencoder architecture where the encoder aims to learn high-level semantic representation across multi-view, while the decoder employs these cross-view fused high-level representation to reconstruct missing view information, guiding the model to fill data gaps.
For another, the given raw data usually come with many redundant features \cite{liu2013global}, and blindly using raw data directly to impute missing values may accumulate data redundancy. To address the challenges, drawing inspiration from the information bottleneck theory \cite{tishby2000information}\cite{tishby2015deep}, we propose a task-oriented feature augmentation network to identify and select a subset of features that are most pertinent to the task, thereby ensuring that clean features are obtained before imputation.

Following the above motivations and considerations, we finally attempt to integrate task-oriented feature augmentation with cross-view imputation for the PMVIMLC task. Simultaneously, to make view-imputation serve the PMVIMLC task, a two-stage training approach is adopted to optimize our network:

$\cdot \mathbf{}$ Stage one --- Task-Oriented Feature Augmentation

$\cdot \mathbf{}$ Stage two --- Cross-View Imputation and Enhanced Classification 

The overall framework is shown in Fig.\ref{figure_one}. The first stage focuses on extracting task-related features from each view. The second stage takes the features obtained from first stage as inputs and recovers missing views to create complete multi-view data. And the network then employs the completed multi-view data to perform classification prediction which fundamentally improve classification accuracy and robustness. 

\subsection{Task-Oriented Feature Augmentation}
When we tackle the complex problem of recovering missing multi-view data, we recognize that original data often contain task-irrelevant redundant information which may undermine the validity of the subsequent imputation process and the final classification results \cite{tan2024two}\cite{tishby2000information}. To this end, we seek to obtain high-quality discriminative and compact representation $V_{l} $ that densely encapsulates task-relevant information to replace the initial view redundant features $X_{l}$ before imputation. This target can be realized by minimizing the following problem according to the information bottleneck principle \cite{kingma2013auto}:
\begin{equation}
\label{eq_one}
L_{ta}^{l} = - I\left ( {V_{l } };Y  \right ) + {\delta  }I\left ( {V_{l } };{X_{l } }  \right )
\end{equation}
where $I\left ( {V_{l } };Y \right )$ represents the mutual information between $l$-th view $V_{l}$ and the real label $Y$, and we aim to maximize it as much as possible, which helps to obtain more task-related information to improve the model's understanding of the task. $I\left ( {V_{l } };{X_{l } }  \right )$ indicates the mutual information between $V_{l}$ and $X_{l}$, and we aspire to minimize it to filter out extraneous information that is irrelevant to the task execution. By doing so, this process can distinguish between useful information and noise so that ensure the model learns task-focused features from different views. And throughout this process, a uniform conditioning factor $\delta $ is introduced to balance the retention of task-relevant information for all views.

However, directly computing the mutual information is often accompanied by high computational complexity. To alleviate this difficulty, firstly, inspired by \cite{tan2024two}\cite{achille2018information}\cite{amjad2019learning}, we approximate the first term $-I\left ( {V_{l } };Y  \right )$ as a generalized common cross entropy loss:
\begin{equation}
	\label{eq_two} 
	\begin{aligned}
-I\left ( {V_{l } };Y  \right )=\frac{1}{nc}\sum_{i=1}^{n} \sum_{j=1}^{c}  \left [ \left ( 1-Y_{i,j}  \right ) \log_{}{\left ( 1- Y^{'} _{l_{i,j} }\right ) }  +Y_{i,j}\log_{}{ Y^{'} _{l_{i,j} }}  \right] 
    \end{aligned}
\end{equation}
where $n$ represents the total number of samples, $c$ represents the total number of labels and $Y^{'} _{l}$ is the output of classifier $\complement _{l}$, i.e., $Y_{l}^{'}=\complement_{l}\left ( V_{l} \right )$. This approximation not only retains a high degree of sensitivity to the task-related information, but also significantly reduces the computational burden and improves the practical feasibility of the algorithm. 

Then, for the second term $I\left ( {V_{l } };{X_{l } }  \right )$, we introduce the variational encoding-decoding technique \cite{bang2021explaining}. And to clarify the subsequent discussion more clearly, we'd better to start by explaining the basic concept of mutual information. That is, given ${V_{l } }$ and ${X_{l } }$, their joint probability distribution is denoted as $p( V_{l } ,X_{l } )$ and their marginal probability distribution are denoted as $p(V_{l })$ and $p(X_{l })$, respectively, then the mutual information $I(V_{l };X_{l })$ is defined as:
\begin{equation}
\label{eq_three}
\begin{aligned}I\left ( V_{l };X_{l } \right )=H\left ( V_{l } \right )-H\left ( V_{l }|X_{l } \right )=\mathbb{E}_{p\left ( V_{l },X_{l } \right ) } \left [ \log_{}{\left ( \frac{p\left ( V_{l },X_{l } \right ) }{p\left ( V_{l } \right ) p\left ( X_{l } \right ) }  \right ) }  \right ]\end{aligned}  
\end{equation}
where $\mathbb{E}$ denotes expectation, $H\left (  V_{l } \right )$ and $H\left (  V_{l }|X_{l } \right )$ represent entropy and conditional entropy, respectively. 

So it is obvious that by bringing the formula $p\left ( V_{l} ,X_{l} \right ) =p\left ( V_{l}|X_{l} \right ) p\left ( X_{l} \right ) $ into Eq.\eqref{eq_three}, then we can get the upper bound:
\begin{equation}
	\label{eq_four} 
	\begin{aligned}
I\left ( V_{l}; X_{l} \right ) \le \mathbb{E}_{p\left ( V_{l},X_{l} \right ) }\left [ \log_{}{\frac{p\left ( V_{l}|X_{l} \right ) }{p\left ( V_{l} \right ) } }  \right ].
\end{aligned}
\end{equation}
And for computational expedience, $p\left ( V_{l} \right )$ can be arbitrarily set as a multivariate Gaussian $N\left ( V_{l} | 0,I\right )$ to be an approximate probability distribution; $p\left ( V_{l}|X_{l} \right )$ is set as another multivariate Gaussian $N\left ( V_{l}|\mu_{l}, {\textstyle \sum_{l}}\right ) $, where both the mean $\left \{ \mu_{l}^{(i)} \subseteq R^{d_{l_{v}}} \right \}_{i=1}^{n}$ and diagonal covariance matrix $\left \{ \textstyle \sum_{l}^{(i)}\subseteq R^{d_{l_{v}}\times d_{l_{v}}}  \right \}_{i=1}^{n}  $ are associated with $X_{l}$. To determine $\mu_{l}$ and ${\textstyle \sum_{l}}$, a specialized multilayer perceptron (MLP) TAEncoder is employed, which accepts the data matrix $X_{l}\in R^{n\times d_{l_{x}}} $ as input and generates two matrices, $\mathcal{P}_{l}\in R^{n\times d_{l_{v}}}$ and $\mathcal{Q}_{l}\in R^{n\times d_{l_{v}}}$, related to the mean and standard deviation of $V_{l}$, respectively. So the generation of $V_{l}$ can be then obtained through $V_{l}=\mathcal{P}_{l}+\varpi \odot \mathcal{Q}_{l} $ \cite{kingma2013auto}. Here,$\varpi$ is a matrix drawn from a standard Gaussian distribution, and $\odot$ denotes the Hadamard product. Subsequently, the two Gaussians $p\left ( V_{l} \right )$ and $p\left ( V_{l}|X_{l} \right )$ are substituted into the upper bound formula Eq.\eqref{eq_four} and a decomposition and simplification step is performed. This step entails not only the splitting of these distributions into multiple one-dimensional Gaussian distributions but also the use of the empirical distributions to approximate the marginal probability distribution $p\left ( X_{l} \right )$. Following this procedure, the second term $I\left ( {V_{l } };{X_{l } }  \right )$ can be reformulated as the following loss:
\begin{equation}
	\label{eq_five}
I\left ( {V_{l } };{X_{l } }  \right )=\frac{1}{2nd_{l_{v} } }\sum_{i=1}^{n}\sum_{j=1}^{d_{l_{v} } }\left[(\mathcal{Q}_{l_{i,j}})^{2}-\log_{}{(\mathcal{Q}_{l_{i,j}})^{2}}+(\mathcal{P}_{l_{i,j}})^{2}-1\right]
\end{equation}
where $d_{l_{v} } $ denotes the feature dimension to the $l$-th view $V_{l}$.

Ultimately, by integrating Eq.\eqref{eq_two} and Eq.\eqref{eq_five}, we establish the final objective function of the optimization in the stage one:
\begin{equation}
	\label{eq_six}
\begin{aligned}	
L_{ta}^{l}&=\frac{1}{nc}\sum_{i=1}^{n}\sum_{j=1}^{c}\left[\left ( 1-Y_{i,j}\right )\log_{}{\left( 1-Y_{l_{i,j} }^{'}\right)}+Y_{i,j}\log_{}{Y_{l_{i,j} }^{'}}\right]U_{i,l}G_{i,j} \\
&+\frac{\delta }{2nd_{l_{v} } }\sum_{i=1}^{n}\sum_{j=1}^{d_{l_{v} } } \left[(\mathcal{Q}_{l_{i,j}})^{2}-\log_{}{(\mathcal{Q}_{l_{i,j}})^{2}}+(\mathcal{P}_{l_{i,j}})^{2}-1\right]U_{i,l}        
\end{aligned}   
\end{equation}
$U$ is the missing view indicator matrix and $G$ is the missing label indicator matrix, which serve to systematically exclude the detrimental interference of missing views and missing labels on the model training process \cite{liu2023dicnet}\cite{ou2024view}.

\subsection{Cross-View Imputation}
In the preceding subsection, we deeply capture the highly task-relevant information across multi-view while filtering out redundant information with the network in stage one. On this basis, we further propose a novel cross-view imputation module to recover missing views yet to enhance the performance.

Different from traditional methods that concentrate on the shallow features of data, we elect deep autoencoders to extract the high-level embedding features of multi-view data. Concretely, we project the task-relevant features obtained in the first stage into the embedding space by view-specific encoders (VSEncoder) consisting of MLPs, ensuring that the extracted high-level embedding features maintain view-specific semantic information, and at the same time making different views in the embedding space aligned with the same dimension to facilitate subsequent feature fusion. The formal expression for the $l$-th view is:  
\begin{equation}
	\label{eq_seven}
Z_{l}=\Psi_{l}\left ( V_{l},\vartheta_{l}\right )    
\end{equation} 
where $Z_{l}\in R^{n\times d_{e}}$ is the extracted embedding features of view $l$, $\Psi_{l}$ is the encoder to $l$-th view and $\vartheta_{l}$ is network parameter corresponding to $\Psi_{l}$.

After obtaining the embedding features of each view using VSEncoder, to fully utilize and effectively complement the unique information of each view, we adopt a straightforward and efficient strategy, namely multi-view weighted fusion \cite{liu2024masked}, which systematically merges all views through weight assignment to form a unified and comprehensive representation as follows:
\begin{equation}
	\label{eq_eight}
\widehat{Z_{i,:}} =\sum_{l=1}^{m} \frac{Z_{l_{i,:} } U_{i,l}}{\sum_{l^{'} }U_{i,l^{'}}} 
\end{equation} 
where $\widehat{Z_{i,:}}\in R^{d_{e}}$ represents the fused representation of the $i$-th sample and $Z_{l_{i,:} } \in R^{d_{e}}$ denotes the discriminant feature of the $i$-th sample of the $l$-th view. The value of $U_{i,l}$ and $U_{i,l^{'}}$ is either be 1 or 0, and $m$ is the total number of views. 

Learning this comprehensive sample representation $\widehat{Z}$ by exploiting the shared property of view-level information described above means that if some views are absent from the sample, a symmetric cross-view decoder module VSDecoder connected to $\widehat{Z}$ can be trained to reconstruct the data for all views, especially targeting the originally missing parts for completion, i.e.:
\begin{equation}
	\label{eq_nine}
V_{l}^{'}=\Phi_{l}\left ( \widehat{Z},\zeta _{l}  \right )      
\end{equation} 
where $V_{l}^{'}\in R^{n\times d_{l_{v} } }$ is the reconstructed features of the $l$-th view, $\Phi_{l}$ is the decoder to $l$-th view and $\zeta _{l}$ is network parameter corresponding to $\Phi_{l}$. Despite the inherent challenges of completely and accurately restoring all missing information, the strategy is effective in inferring predicted values close to the real data which is attributed to the inherent cross-view similar knowledge of the shared representation $\widehat{Z}$. And given that we do not have direct access to supervisory information targeting the accuracy of the reconstructed data, we adopt an alternative strategy that drawing on the available initial data to impose a constraint known as the weighted reconstruction loss \cite{liu2024masked}:
\begin{equation}
	\label{eq_ten}
L_{re}=\frac{1}{nm}\sum_{i=1}^{n}\sum_{l=1}^{m}\frac{1} {d_{l_{v} }}  \parallel V_{l_{i,:}}^{'}- V_{l_{i,:}}\parallel _{2}^{2}U_{i,l}            
\end{equation} 
where $V_{l_{i,:}}^{'}$ is the reconstructed instance corresponding to the $l$-th view of the $i$-th sample, i.e.,$V_{l_{i,:}}$. By doing so, despite the lack of direct discriminative guidance, we are able to lead the model to focus on the retention and reconstruction of critical information, thus indirectly facilitating the learning process \cite{liu2023information}. To the end, as illustrated in Fig.\ref{figure_one}, by inserting the features obtained by decoding and reconstruction into the missing view positions in the original sample, a complete view set $\left \{ T_{l} \right \} _{l=1}^{m}$ is obtained, which enjoys more information and therefore cross-view consistency and complementarity can be better exploited to facilitate the final classification.

\subsection{Incomplete Multi-Label Classification}
Once the complete views $\left \{ T_{l} \right \} _{l=1}^{m}$ have been obtained, they are then used for the final, incomplete multi-label classification task. Firstly, view-specific encoders $\left \{ \Psi_{l} \right \} _{l=1}^{m}$ are employed to capture the deep semantic information in each $T_{l}$:
\begin{equation}
	\label{eq_eleven}
Z_{T}^{l} =\Psi_{l}\left ( T_{l} ,\lambda_{l} \right )   
\end{equation} 
where $Z_{T}^{l}\in R^{n\times d_{e}}$ is the high-level features of $T_{l}$, and $\lambda_{l}$ is network parameter corresponding to $\Psi_{l}$.

Then, our target is to create a unified and consistent representation which adequately characterizes multi-view data. However, different views may make varying contributions to such a representation. To account for the different discriminative capabilities of different views, and the relative values explorability of individual view under the data completeness enhancement, we transition to a weighted fusion approach with learnable weights as follow when the originally missing views are made complete to substitute the fixed-weight fusion method employed when multi-view data are incomplete \cite{trosten2021reconsidering}:
\begin{equation}
	\label{eq_twelve}
\widehat{Z_{T_{i,:}}} =\frac{1}{m} \sum_{l=1}^{m}\sigma_{l}  Z_{T_{i,:}}^{l} 
\end{equation}  
$\widehat{Z_{T_{i,:}}}\in R^{d_{e}}$ represents the fused representation of the $i$-th sample to $\widehat{Z_{T}}\in R^{n\times d_{e}}$ and $\widehat{Z_{T}}$ serves as an input to the next classification layer. $Z_{T_{i,:}}^{l}  \in R^{d_{e}}$ denotes the discriminant features of the $i$-th sample to $l$-th view and $\left \{ \sigma_{l} \right \} _{l=1}^{m} $ refer to learnable weights processed by softmax. 
  
Finally, we construct a direct mapping from the joint feature space to the multi-label classification space. It is achieved by applying two linear layers with a sigmoid activation function that independently generates a probability score for each potential category in each sample. These probability scores are bounded by a sigmoid transformation between 0 and 1. Mathematically, this classification process can be abstracted as $\left \{  \mathbb{C} :\widehat{Z_{T}}\in R^{n\times d_{e} } \longrightarrow Y^{'}\in \left [ 0,1 \right ] ^{n\times c}  \right \}$, where $Y^{'}$ represents the probability distribution of $c$ categories predicted for $n$ samples, $\mathbb{C}$ denotes the classifier. And to address the issue of missing labels, we employ a missing label indicator matrix, denoted by $G$, to adjust the BCE loss function \cite{liu2022incomplete}\cite{liu2024attention}. This result in the proposal of a weighted binary cross-entropy loss function, which assesses the difference between the predicted label $Y^{'}$ and the real label $Y$. This assessment guides the model training by following:
\begin{equation}
	\label{eq_thirteen}
	\begin{aligned}
L_{c}=-\frac{1}{nc}\sum_{i= 1}^{n}\sum_{j = 1}^{c}\left [ Y_{i,j}\log_{}{Y_{i,j}^{'} }+\left ( 1-Y_{i,j} \right )\log_{}{\left ( 1-Y_{i,j}^{'}  \right ) } \right ]G_{i,j} 
     \end{aligned}     
\end{equation} 

Combining Eq.\eqref{eq_ten} with Eq.\eqref{eq_thirteen}, the overall objective function for stage two can be obtained:
\begin{equation}
	\label{eq_fourteen}
	L_{fc}=L_{c}+\alpha L_{re} 
\end{equation}
where $\alpha$ indicates penalty coefficient for $L_{re}$ and it is used to balance the relative importance between classification loss $L_{c}$ and reconstruction loss $L_{re}$.

\begin{algorithm}[!t]
	\caption{Training process of TACVI-Net}\label{algorithm1}
	\begin{algorithmic}
		\State 
		\State \textbf{Input:} Partial multi-view data $\left \{ X_{l} \right \} _{l=1}^{m}$, accompanied by a missing-view indicator matrix $U\in\left \{ 0,1 \right \} ^{n\times m} $; Incomplete multi-label matrix $Y\in R^{n\times c} $, accompanied by a missing-label indication matrix $G\in\left \{ 0,1 \right \} ^{n\times c}$; batch size $B$; maximum number of iterations $T$; learning rate $l_{r}$; hyper-parameters $\alpha$ and $\left \{ \delta _{l}  \right \} _{l=1}^{m} $.
		\State \textbf{Initialization:} Fill in missing elements with 0, initialize TAEncoder $\left \{ \backepsilon_{l} \right \} _{l=1}^{m}$, VSEncoder $\left \{ \Psi_{l} \right \} _{l=1}^{m}$, VSDecoder $\left \{ \Phi_{l} \right \} _{l=1}^{m}$, classifiers $\left \{ \complement_{l} \right \} _{l=1}^{m}$ and $\mathbb{C}$.
		\State \textbf{procedure} Stage one
		\For{$l=1$ to $m$} 
		  \State calculate $V_{l}=\backepsilon_{l}\left ( X_{l}\right )$, $Y_{l}^{'} =\complement _{l} \left ( V_{l}\right )$;
		\State calculate $L_{ta}^{l}$ by Eq.\eqref{eq_six};
		\State update $\backepsilon_{l}$ and $\complement _{l}$;
		\EndFor	
		\State \textbf{end procedure}
		\State \textbf{procedure} Stage two
		\For{$t=1$ to $T$}
		  \State calculate multi-view embedding feature $Z_{l}$ by Eq.\eqref{eq_seven};
		 \State calculate fusion feature $\widehat{Z}$ by Eq.\eqref{eq_eight};
		 \State calculate reconstructed feature $V_{l}^{'}$ by Eq.\eqref{eq_nine};
		 \State calculate reconstructed loss $L_{re}$ by Eq.\eqref{eq_ten}; 
		 \State complete missing data in $X_{l}$ with $V_{l}^{'}$ to get new data $T_{l}$;
		 \State calculate multi-view embedding feature $Z_{T}^{l}$ by Eq.\eqref{eq_eleven};
		 \State calculate fusion feature $\widehat{Z_{T}}$ by Eq.\eqref{eq_twelve};
		 \State calculate classification loss $L_{c}$ by Eq.\eqref{eq_thirteen};
		 \State update $\Psi_{l}$, $\Phi_{l}$, $\mathbb{C}$ and $\sigma_{l}$;
		\EndFor 	
		\State \textbf{end procedure}
		\State \textbf{Output:} Trained model
	\end{algorithmic}
	\label{alg1}
\end{algorithm}

\section{Experiment}
In this segment, we present the configuration of the dataset, the methods of comparison, and provide a detailed analysis of the results from the experiment.

\subsection{Experimental Settings}

\begin{table}
	\centering
	\caption{Statistics of five multi-view multi-label datasets.}
	\label{table_two}
        \resizebox{0.45\textwidth}{!}{
	\begin{tabular}{c @{\hspace{25pt}} c @{\hspace{25pt}} c @{\hspace{25pt}} c}
		\toprule
		dataset & \#sample & \#label & \#label/\#sample \\
		\midrule  
		Corel5k & 4999 & 260 & 3.396 \\
		Pascal07 & 9963 & 20 & 1.465 \\
		ESPGame & 20770 & 268 & 4.686 \\
		IAPRTC12 & 19627 & 291 & 5.719 \\
		MIRFLICKR & 25000 & 38 & 4.716 \\
		\bottomrule
	\end{tabular}
        }
\end{table}

\subsubsection{Datasets}
In accordance with previous studies \cite{liu2023dicnet}\cite{tan2018incomplete}\cite{li2021concise}, we have chosen five widely used multi-view multi-label datasets to verify our proposed model. These datasets include Corel5k, Pascalo7, ESPGame, IAPRTC12, and MIRFLICKR. Each dataset encompasses six views, specifically GIST, HSV, HUE, SIFT, RGB, and LAB. A brief summary of these datasets can be found in Table \ref{table_two}, with further detailed descriptions provided as follows:

\textbf{Corel5k} \cite{duygulu2002object}: Corel5k is comprised of 4,999 images that have been annotated with 260 categories. It should be noted that each image is affiliated with at least one of these categories.

\textbf{Pascal07} \cite{everingham2007pascal}: Pascal07 is a widely utilized resource in the field of visual object detection and recognition, comprising 9,963 images and 20 categories of objects, many of which contain multiple instances of an object.

\textbf{ESPGame} \cite{von2004labeling}: ESPGame comprises 20,770 images of the game, each of which has been annotated with one or more labels.

\textbf{IAPRTC12} \cite{grubinger2006iapr}: IAPRTC12 comprises a substantial corpus of 19,627 image samples and 291 categories.

\textbf{MIRFLICKR} \cite{huiskes2008mir}: MIRFLICKR consists of 25,000 images sourced from the Flickr website, which employs a total of 38 distinct tags.\\

\subsubsection{Incomplete Multi-view Multi-label Data Preparation}
Given that all five datasets mentioned above have complete views and labels, they cannot be used directly to evaluate the efficacy of our model in situations with missing data. Following the methodology in \cite{liu2023dicnet}\cite{tan2018incomplete}\cite{li2021concise}, we preprocess the complete dataset into a partial multi-view incomplete multi-label dataset. More specifically, we establish samples with a randomly missing percentage, denoted as $a$\%, in each view, ensuring that each sample is present in at least one view, thereby creating a partial multi-view dataset. We then randomly eliminate $b$\% of both positive and negative labels for each category to simulate a scenario with missing labels. Finally, the dataset is randomly divided into three subsets: 70\% for training, 15\% for validation, and 15\% for testing. Note that any missing element in views and labels is substituted with 0. \\

\subsubsection{Competitors}
In the experimental session, we compare our TACVI-Net with four classical models and five recent popular approaches: \textbf{GLOCAL} \cite{zhu2017multi}, \textbf{CDMM} \cite{zhao2021consistency}, \textbf{iMvML} \cite{tan2018incomplete}, \textbf{NAIM3L} \cite{li2021concise}, \textbf{DICNet} \cite{liu2023dicnet}, \textbf{AIMNet} \cite{liu2024attention}, \textbf{MTD} \cite{liu2024masked}, \textbf{SIP} \cite{liupartial}, and \textbf{ATSINet} \cite{tan2024two}. Since GLOCAL and AIMNet have not been discussed in detail in Section \ref{relatedworks}, a brief description of these two methods is necessary here. GLOCAL emphasizes the exploration of label correlations from both global and local views. AIMNet is an attention-induced missing instance filling technique based on an attempt to approximate potential features of missing instances in the embedding space through cross-view joint attention. To address the incomplete multi-label classification property of GLOCAL, which is only applicable to single views, we conduct experiments independently for each view and select the best result to characterize the multi-view classification performance. In addition, as CDMM lacks the capacity to address missing data, we make suitable corrections to the model by implementing a mean-filling strategy. Our experiment is conducted ten times repetitively, with both the mean and standard deviation of the results being documented. And the final experimental results are obtained by selecting the model parameters that performed best on the validation set during the optimization process and evaluating them on the test set.\\ 

\subsubsection{Evaluation metrics}
Consistent with prior studies \cite{liu2023dicnet}\cite{tan2018incomplete}\cite{li2021concise}, we employ four commonly utilized evaluation metrics for multi-label classification to assess our model's performance, specifically: Average Precision (AP), Hamming Loss (HL), Ranking Loss (RL), and Area Under the Receiver Operating Characteristic Curve (AUC). To render the metrics more intuitively illustrative of the model's performance, we convert HL into its inverse (1-HL) and similarly invert RL to (1-RL), where a higher score for each metric denotes superior performance.\\

\subsubsection{Implementation details}
Our TACVI-Net model is implemented utilizing Python programming language and the Pytorch framework. For model training, we employ the Stochastic Gradient Descent (SGD) optimizer, with a momentum value set to 0.9. The configuration includes a batch size of 128 and a learning rate of 0.0001. The computational environment for running our model comprises an Ubuntu operating system, equipped with an RTX3080Ti GPU.

\begin{table*}
	\centering
	\caption{The experimental results of ten methods on five datasets with 50\% missing views, 50\% missing labels, and 70\% training samples. The higher the value, the better the performance. Bold indicates both the best and the suboptimal results.}
	\label{tabel_three}
	\resizebox{0.99\textwidth}{!}{
		\begin{tabular}{cccccccccccc}
			\toprule
			DATA & METRIC & GLOCAL \cite{zhu2017multi} & CDMM \cite{zhao2021consistency}  & iMvWL \cite{tan2018incomplete} & NAIM3L \cite{li2021concise} & DICNet \cite{liu2023dicnet} & AIMNet \cite{liu2024attention} & MTD \cite{liu2024masked} & SIP \cite{liupartial} & ATSINet \cite{tan2024two} & TACVI-Net\\
			\midrule
			& AP & 0.285$\pm$0.004 & 0.354$\pm$0.004 & 0.283$\pm$0.008 & 0.309$\pm$0.004 & 0.381$\pm$0.004 & 0.400$\pm$0.010 & 0.415$\pm$0.008 & 0.418$\pm$0.009 & \uline{0.436$\pm$0.007} & \textbf{0.447$\pm$0.011} \\
			
			& 1-HL & \uline{0.987$\pm$0.000} & \uline{0.987$\pm$0.000} & 0.978$\pm$0.000 & \uline{0.987$\pm$0.000} & \textbf{0.988$\pm$0.000}  & \textbf{0.988$\pm$0.000} & \textbf{0.988$\pm$0.000} & \textbf{0.988$\pm$0.000} & \textbf{0.988$\pm$0.000} &  \textbf{0.988$\pm$0.000} \\
			
			Corel5k & 1-RL & 0.840$\pm$0.003 & 0.884$\pm$0.003 & 0.865$\pm$0.005 & 0.878$\pm$0.002 & 0.882$\pm$0.004  & 0.902$\pm$0.002 & 0.893$\pm$0.004 & 0.911$\pm$0.003 & \uline{0.917$\pm$0.002} & \textbf{0.921$\pm$0.002}\\
			
			& AUC &  0.843$\pm$0.003 &  0.888$\pm$0.003 &  0.868$\pm$0.005 &  0.881$\pm$0.002 &  0.884$\pm$0.004  & 0.905$\pm$0.003 & 0.896$\pm$0.004 & 0.913$\pm$0.003 & \uline{0.920$\pm$0.002} &  \textbf{0.923$\pm$0.002} \\		
			\midrule
			& AP & 0.496$\pm$0.004 & 0.508$\pm$0.005 & 0.437$\pm$0.018 & 0.488$\pm$0.003 & 0.505$\pm$0.012  & 0.548$\pm$0.008 & 0.551$\pm$0.004 & 0.555$\pm$0.010 & \uline{0.581$\pm$0.009} &  \textbf{0.588$\pm$0.009} \\
			
			& 1-HL & 0.927$\pm$0.000 & 0.931$\pm$0.001 & 0.882$\pm$0.004 & 0.928$\pm$0.001 & 0.929$\pm$0.001  & 0.931$\pm$0.001 & 0.932$\pm$0.001 & 0.931$\pm$0.001 & \uline{0.934$\pm$0.001} &  \textbf{0.935$\pm$0.001} \\
			
			Pascal07 & 1-RL & 0.767$\pm$0.004 & 0.812$\pm$0.004 & 0.736$\pm$0.015 & 0.783$\pm$0.001 & 0.783$\pm$0.008  & 0.831$\pm$0.004 & 0.831$\pm$0.003 & 0.830$\pm$0.004 & \uline{0.849$\pm$0.005} & \textbf{0.853$\pm$0.005}\\
			
			& AUC &  0.786$\pm$0.003 &  0.838$\pm$0.003 &  0.767$\pm$0.015 &  0.811$\pm$0.001 &  0.809$\pm$0.006  & 0.851$\pm$0.004 & 0.851$\pm$0.003 & 0.850$\pm$0.005 & \uline{0.868$\pm$0.004} &  \textbf{0.871$\pm$0.004} \\
			\midrule
			& AP & 0.221$\pm$0.002 & 0.289$\pm$0.003 & 0.244$\pm$0.005 & 0.246$\pm$0.002 & 0.297$\pm$0.002  & 0.305$\pm$0.004 & 0.306$\pm$0.003 & 0.311$\pm$0.004 & \uline{0.319$\pm$0.004} & \textbf{0.334$\pm$0.005} \\
			
			& 1-HL & \uline{0.982$\pm$0.000} & \textbf{0.983$\pm$0.000} & 0.972$\pm$0.000 & \textbf{0.983$\pm$0.000} & \textbf{0.983$\pm$0.000}  & \textbf{0.983$\pm$0.000} & \textbf{0.983$\pm$0.000} & \textbf{0.983$\pm$0.000} & \textbf{0.983$\pm$0.000} &  \textbf{0.983$\pm$0.000} \\
			
			ESPGame & 1-RL & 0.780$\pm$0.004 & 0.832$\pm$0.001 & 0.808$\pm$0.002 & 0.818$\pm$0.002 & 0.832$\pm$0.001  & 0.846$\pm$0.002 & 0.837$\pm$0.002 & 0.849$\pm$0.002 & \uline{0.859$\pm$0.002} & \textbf{0.863$\pm$0.002}\\
			
			& AUC &  0.784$\pm$0.004 &  0.836$\pm$0.001 &  0.813$\pm$0.002 &  0.824$\pm$0.002 &  0.836$\pm$0.001  & 0.850$\pm$0.002 & 0.842$\pm$0.002 & 0.853$\pm$0.002 & \uline{0.863$\pm$0.002} &  \textbf{0.866$\pm$0.001} \\
			\midrule
			& AP & 0.256$\pm$0.002 & 0.305$\pm$0.004 & 0.237$\pm$0.003 & 0.261$\pm$0.001 & 0.323$\pm$0.001  & 0.329$\pm$0.005 & 0.332$\pm$0.003 & 0.331$\pm$0.006 & \uline{0.361$\pm$0.004} & \textbf{0.368$\pm$0.005} \\
			
			& 1-HL & \uline{0.980$\pm$0.000} & \textbf{0.981$\pm$0.000} & 0.969$\pm$0.000 & \uline{0.980$\pm$0.000} & \textbf{0.981$\pm$0.000}  & \textbf{0.981$\pm$0.000} & \textbf{0.981$\pm$0.000} & \uline{0.980$\pm$0.000} & \textbf{0.981$\pm$0.000} &  \textbf{0.981$\pm$0.000} \\
			
			IAPRTC12 & 1-RL & 0.825$\pm$0.002 & 0.862$\pm$0.002 & 0.833$\pm$0.002 & 0.848$\pm$0.001 & 0.873$\pm$0.001 & 0.883$\pm$0.003 & 0.875$\pm$0.002 & 0.885$\pm$0.003 & \uline{0.898$\pm$0.003} & \textbf{0.901$\pm$0.002}\\
			
			& AUC &   0.830$\pm$0.001 &  0.864$\pm$0.002 &  0.835$\pm$0.001 &  0.850$\pm$0.001 &  0.874$\pm$0.000  & 0.885$\pm$0.003 & 0.876$\pm$0.001 & 0.886$\pm$0.002 & \uline{0.899$\pm$0.002} &  \textbf{0.901$\pm$0.002} \\
			
			\midrule
			& AP &  0.537$\pm$0.002 & 0.570$\pm$0.002 &  0.490$\pm$0.012 & 0.551$\pm$0.002 & 0.589$\pm$0.005  & 0.602$\pm$0.004 & 0.607$\pm$0.004 & 0.614$\pm$0.004 & \uline{0.631$\pm$0.003} &  \textbf{0.633$\pm$0.003} \\
			
			& 1-HL & 0.874$\pm$0.001 & 0.886$\pm$0.001 &  0.839$\pm$0.002 & 0.882$\pm$0.001 & 0.888$\pm$0.002  & 0.890$\pm$0.001 & \uline{0.891$\pm$0.001} & \uline{0.891$\pm$0.001} & \textbf{0.895$\pm$0.001} &  \textbf{0.895$\pm$0.001} \\
			
			MIRFLICKR & 1-RL & 0.832$\pm$0.001 & 0.856$\pm$0.001 &  0.803$\pm$0.008 & 0.844$\pm$0.001 & 0.863$\pm$0.004  & 0.873$\pm$0.002 & 0.875$\pm$0.002 & 0.877$\pm$0.002 & \uline{0.887$\pm$0.001} & \textbf{0.888$\pm$0.001}\\
			
			& AUC &  0.828$\pm$0.001 &  0.846$\pm$0.001 &   0.787$\pm$0.012 &  0.837$\pm$0.001 & 0.849$\pm$0.004 &  0.861$\pm$0.001  & 0.862$\pm$0.002 & 0.860$\pm$0.003 & \uline{0.872$\pm$0.003} &  \textbf{0.873$\pm$0.002} \\
			\bottomrule
		\end{tabular}
	}
\end{table*}

\begin{figure}[!t]
	\centering
	\subfloat[missing-view ratios]{\includegraphics[width=1.9in,height=1.3in]{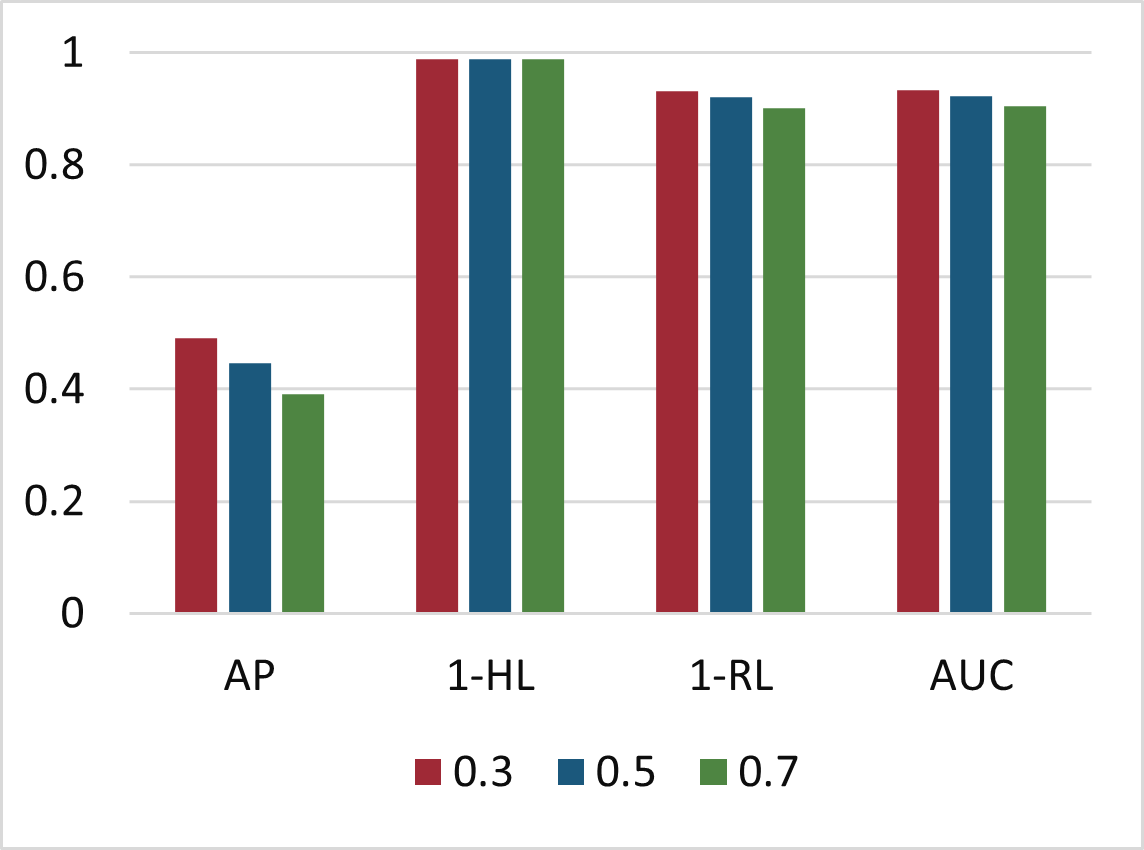}}
	\hfil
	\subfloat[missing-label ratios]{\includegraphics[width=1.9in,height=1.3in]{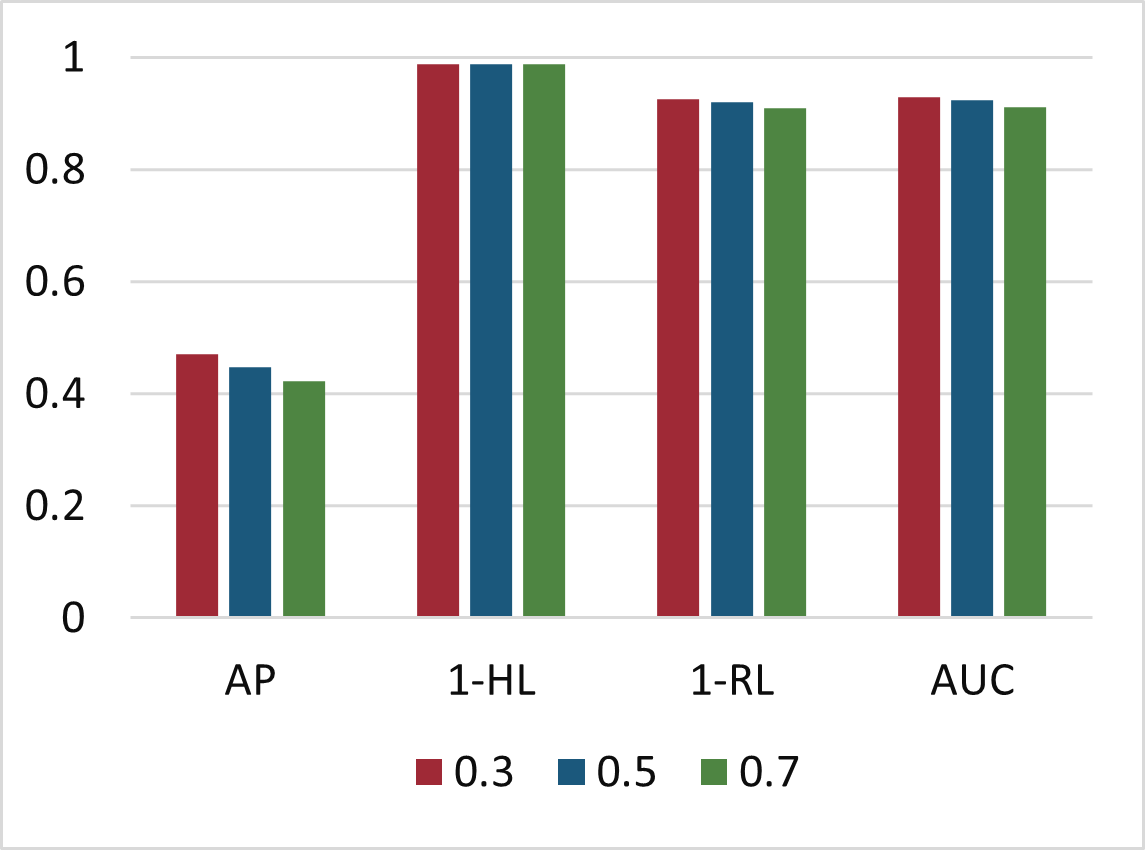}}
	\caption{The performance outcomes for Corel5k are depicted under two scenarios: (a) various proportions of missing views accompanied by a 50\% rate of missing labels, and (b) a steady 50\% rate of missing views coupled with differing levels of missing labels.}
	\label{figure_two}
\end{figure}

\begin{figure}[!t]
	\centering
        \subfloat[30\% view-missing ratio and 70\% 
    label-missing ratio]{\includegraphics[width=1.9in,height=1.3in]{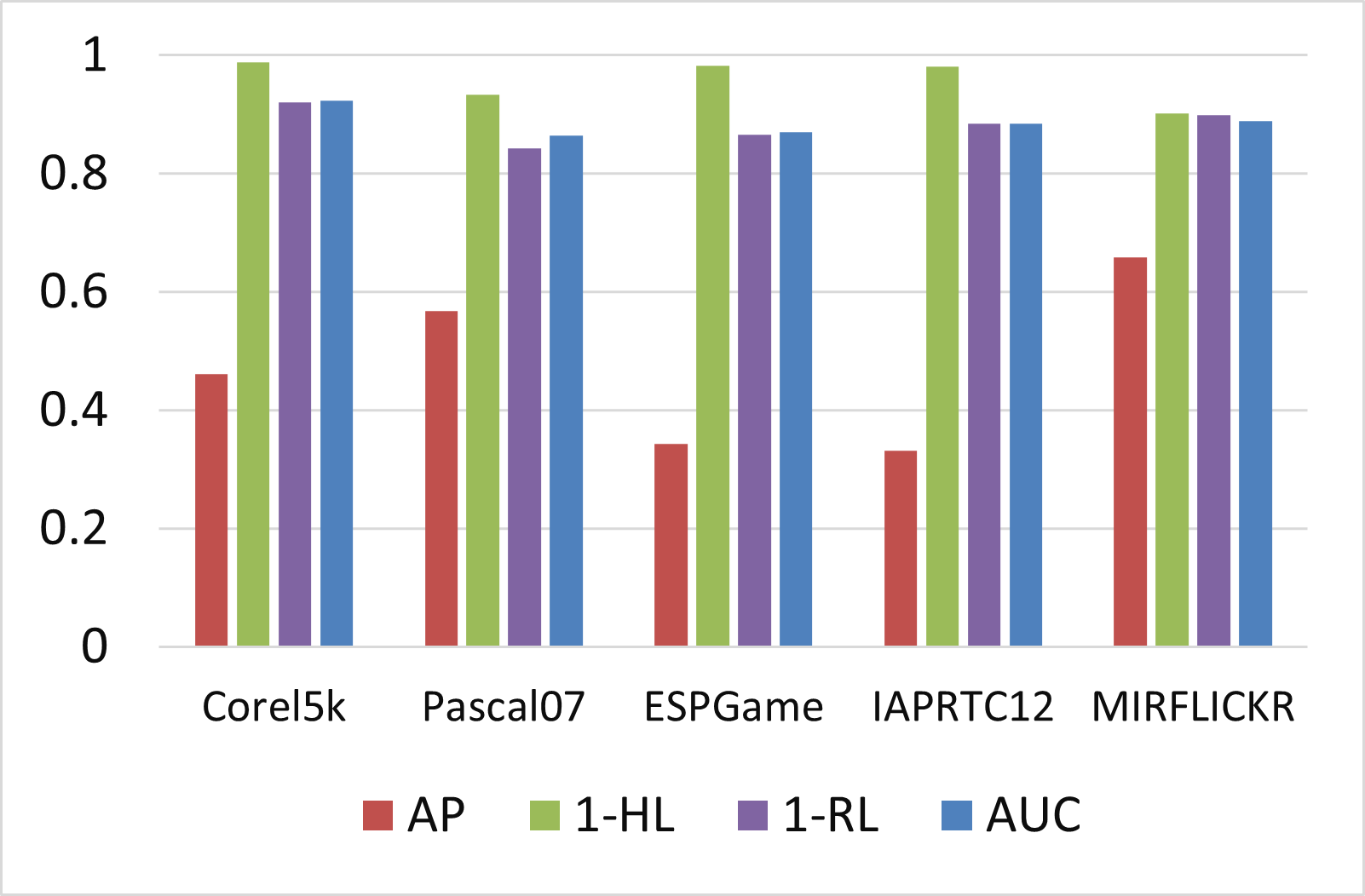}}
    \hfil
        \subfloat[70\% view-missing ratio and 30\% 
  label-missing ratio]
  {\includegraphics[width=1.9in,height=1.3in]{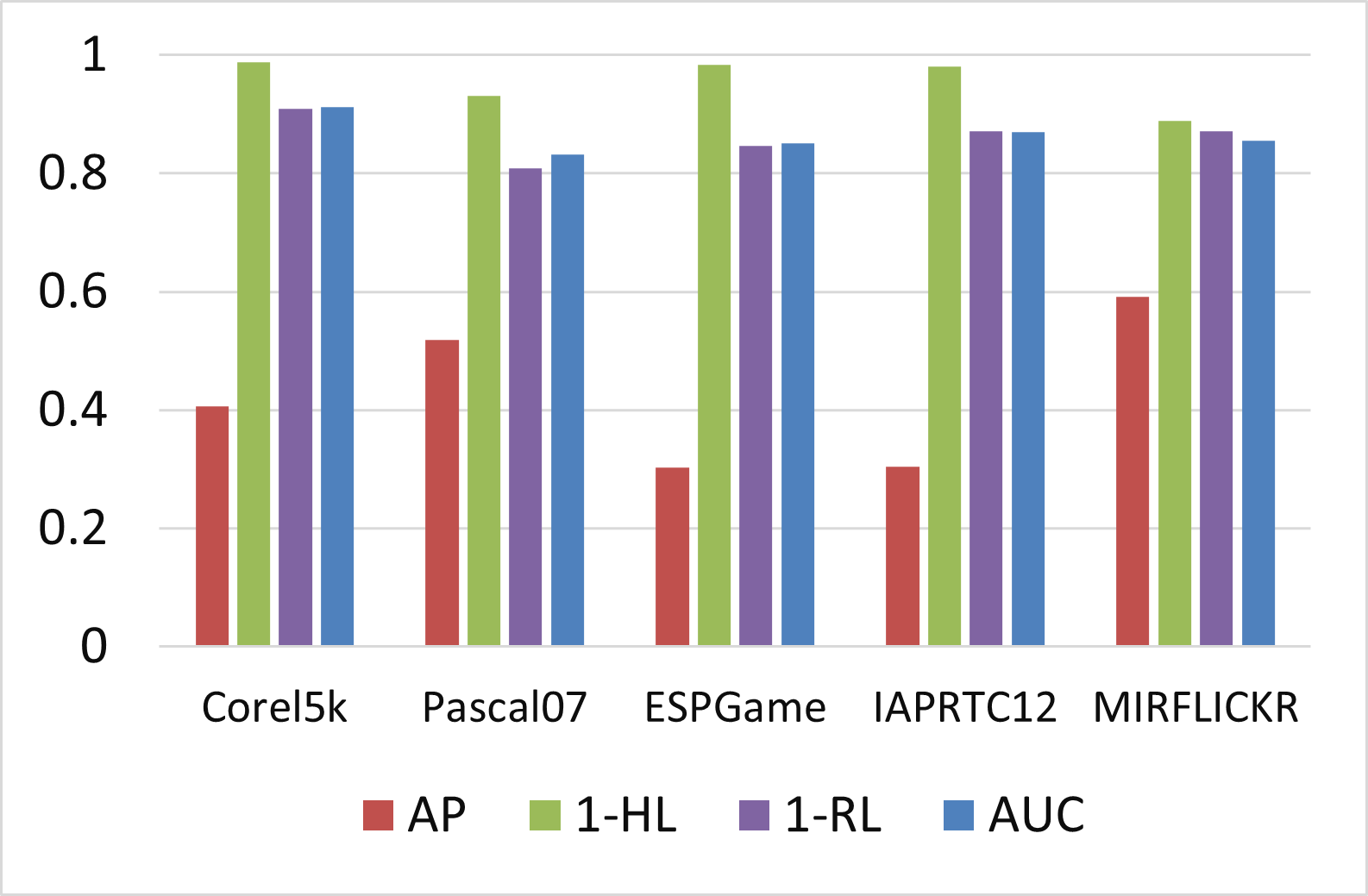}}
	\caption{More experimental results for all datasets with different view-missing ratios and different label-missing ratios.}
	\label{figure_three}
\end{figure}

\begin{figure}[!t]
	\centering
\subfloat[Corel5k]{\includegraphics[width=1.9in,height=1.3in]{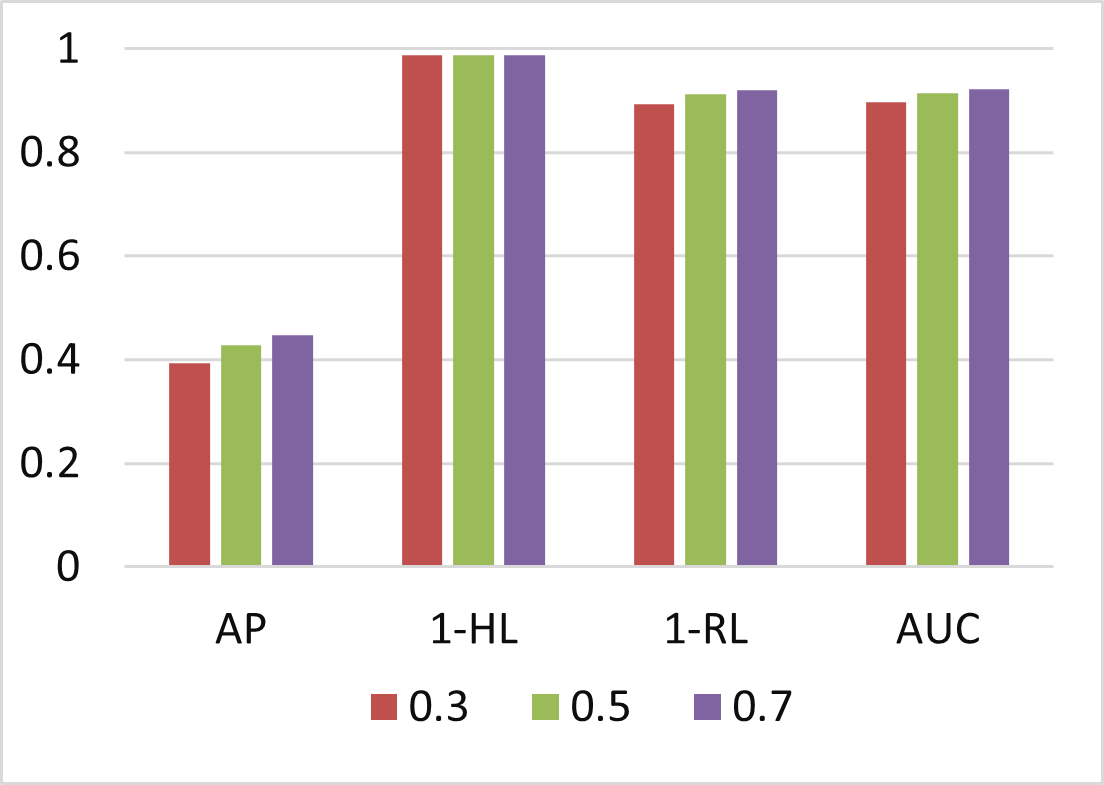}}
\hfil
\subfloat[Pascal07]{\includegraphics[width=1.9in,height=1.3in]{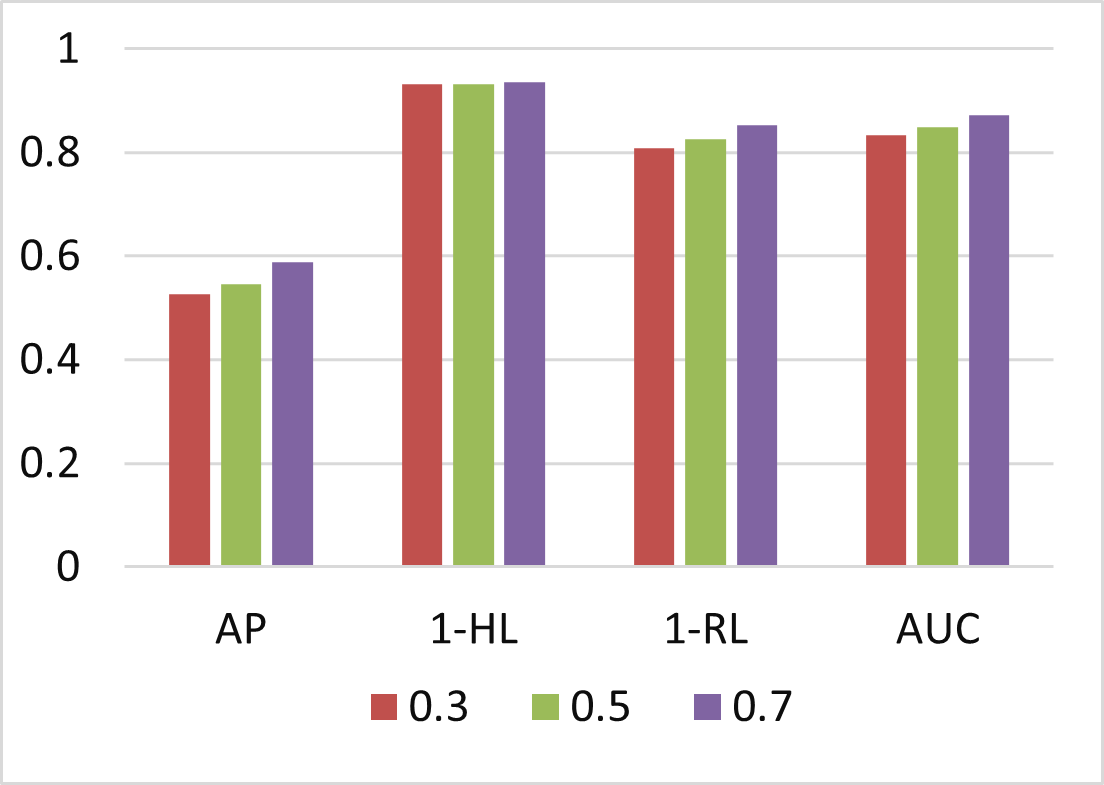}}
	\caption{The performance outcomes of different training sample ratios on (a) Corel5k dataset and (b) Pascal07 dataset with 50\% missing-view ratio and 50\% missing-label ratio.}
	\label{figure_four}
\end{figure}

\subsection{Experimental Result and Analysis}
In Table \ref{tabel_three}, we present the experimental outcomes for ten different methods applied to five datasets, under conditions where 50\% of instances are missing, 50\% of labels are absent, and only 70\% of the samples are utilized for training. And the standard deviation is indicated by the numeral in the lower right corner of each reported result. To investigate the effect of varying levels of missing data in terms of views and labels on our model's performance, we maintain one missing rate at 50\% while adjusting the other missing rate to a series comprising 30\%, 50\%, and 70\%. The resultant outcomes are documented in Fig.\ref{figure_two}. In addition, we extend our experiments to datasets with view and label missing rates both of 30\%, 50\% and 70\%, respectively, as shown in Fig.\ref{figure_three}. From the analysis of Table \ref{tabel_three}, Fig.\ref{figure_two}, Fig.\ref{figure_three} and Fig.\ref{figure_four}, several key observations can be readily deduced:

\begin{itemize}
	\item[$\bullet$] Based on the consolidated analysis presented in Table \ref{tabel_three}, our proposed methodology significantly outperforms all current approaches on the majority of selected evaluation metrics, validating its effectiveness and marking a remarkable lead. Furthermore, techniques such as DICNet and ATSINet demonstrate superior performance compared to the first four methodologies, thanks to their carefully designed deep learning architectures that are explicitly tailored to tackle the intricacies of the PMVIMLC task. This observation underscores the exceptional potential of deep learning in unraveling the inherent complexities associated with this specific challenge, and highlights its ability to tackle such complicated problems.
	\item[$\bullet$] The graphical representation depicted in Fig.\ref{figure_two} and Fig.\ref{figure_three} provides compelling evidence that the absence of complete data views and the presence of incomplete labels within the dataset exert a notably negative influence on the holistic performance metrics of the model. This adverse impact escalates dramatically as the magnitude of missing data within the dataset rises, painting a stark picture of the model's vulnerability to data incompleteness. Most importantly, the study uncovers a key insight: even when the percentage of missing data is constant, the primary obstacle to model effectiveness arises from the inadequacy of view information, surpassing the challenge posed by partially missing labels. This revelation underscores the urgency and importance of employing effective missing view recovery strategies to address such challenges.
        \item[$\bullet$] Fig.\ref{figure_four} illustrates the model's 
    performance across different proportions of training samples. Although the influence of these ratios on the model's efficacy is not overwhelmingly pronounced, a prevailing tendency emerges: the model's performance tends to enhance as the training sample ratio escalates. Consequently, adopting a 70\% allocation for training samples in our experiments is deemed a suitable decision.
\end{itemize}

Although our TACVI-Net chiefly targets the challenge of PMVIMLC, we extend our evaluation to encompass the model's performance within a full-view and full-label setting. The results of this assessment are presented in Table \ref{table_four}. Notably, the model retains its superior performance under MVMLC, indicating that our approach is not only efficacious for PMVIMLC but also holds applicability within MVMLC.

\begin{table*}
    \centering
    \caption{The experimental results of different methods on five data sets with full views and labels. The higher the value, the better the performance. Bold indicates the best result.}
    \label{table_four}
    \resizebox{0.99\textwidth}{!}{
        \begin{tabular}{cccccccccccc}
            \toprule
            DATA & METRIC & GLOCAL \cite{zhu2017multi} & CDMM \cite{zhao2021consistency}  & iMvWL \cite{tan2018incomplete} & NAIM3L \cite{li2021concise} & DICNet \cite{liu2023dicnet} & MTD \cite{liu2024masked} & ATSINet \cite{tan2024two} & TACVI-Net\\
            \midrule
            & AP & 0.386$\pm$0.006 & 0.489$\pm$0.004 & 0.313$\pm$0.005 & 0.327$\pm$0.004 & 0.509$\pm$0.002 & 0.523$\pm$0.003 & \uline{0.543$\pm$0.012} & \textbf{0.567$\pm$0.011} \\
            
            & 1-HL & 0.987$\pm$0.000 & 0.988$\pm$0.000 & 0.979$\pm$0.000 & 0.987$\pm$0.000 & \uline{0.989$\pm$0.000}  & \uline{0.989$\pm$0.000} & \uline{0.989$\pm$0.000} &  \textbf{0.990$\pm$0.000} \\
            
            Corel5k & 1-RL & 0.891$\pm$0.002 & 0.926$\pm$0.001 & 0.884$\pm$0.002 & 0.890$\pm$0.002 & 0.929$\pm$0.001  & 0.931$\pm$0.001 & \uline{0.949$\pm$0.003} & \textbf{0.953$\pm$0.002}\\
            
            & AUC &  0.895$\pm$0.002 &  0.929$\pm$0.001 &  0.887$\pm$0.002 &  0.893$\pm$0.002 &  0.931$\pm$0.000  & 0.934$\pm$0.001 & \uline{0.951$\pm$0.002} &  \textbf{0.955$\pm$0.002}\\        
            \midrule
            & AP & 0.610$\pm$0.005 & 0.589$\pm$0.006 & 0.468$\pm$0.004 & 0.496$\pm$0.003 & 0.608$\pm$0.003  & 0.648$\pm$0.006 & \uline{0.681$\pm$0.007} &  \textbf{0.698$\pm$0.009} \\
            
            & 1-HL & 0.926$\pm$0.000 & 0.931$\pm$0.001 & 0.888$\pm$0.001 & 0.929$\pm$0.001 & 0.937$\pm$0.001  & 0.940$\pm$0.001 & \uline{0.943$\pm$0.006} &  \textbf{0.944$\pm$0.001} \\
            
            Pascal07 & 1-RL & 0.866$\pm$0.002 & 0.873$\pm$0.003 & 0.763$\pm$0.007 & 0.795$\pm$0.001 & 0.859$\pm$0.003  & 0.886$\pm$0.003 & \uline{0.904$\pm$0.004} & \textbf{0.910$\pm$0.005}\\
            
            & AUC &  0.879$\pm$0.002 &  0.893$\pm$0.002 &  0.793$\pm$0.005 &  0.822$\pm$0.001 &  0.876$\pm$0.002  & 0.897$\pm$0.003 & \uline{0.914$\pm$0.004} &  \textbf{0.918$\pm$0.004} \\
            \midrule
            & AP & 0.264$\pm$0.002 & 0.392$\pm$0.002 & 0.260$\pm$0.004 & 0.251$\pm$0.002 & 0.391$\pm$0.001  & \uline{0.397$\pm$0.002} & 0.373$\pm$0.005 & \textbf{0.410$\pm$0.005} \\
            
            & 1-HL & \uline{0.983$\pm$0.000} & \uline{0.983$\pm$0.000} & 0.972$\pm$0.000 & \uline{0.983$\pm$0.000} & \textbf{0.984$\pm$0.000}  & \textbf{0.984$\pm$0.000} & \uline{0.983$\pm$0.000} &  \textbf{0.984$\pm$0.000} \\
            
            ESPGame & 1-RL & 0.804$\pm$0.003 & 0.875$\pm$0.001 & 0.817$\pm$0.001 & 0.825$\pm$0.001 & 0.871$\pm$0.003  & 0.880$\pm$0.001 & \uline{0.888$\pm$0.002} & \textbf{0.898$\pm$0.002}\\
            
            & AUC &  0.810$\pm$0.003 &  0.878$\pm$0.001 &  0.822$\pm$0.001 &  0.830$\pm$0.002 &  0.874$\pm$0.004  & 0.883$\pm$0.001 & \uline{0.891$\pm$0.001} &  \textbf{0.901$\pm$0.001} \\
\midrule
            & AP & 0.330$\pm$0.003 & 0.424$\pm$0.003 & 0.250$\pm$0.004 & 0.267$\pm$0.001 & 0.418$\pm$0.006  & 0.443$\pm$0.003 & \uline{0.451$\pm$0.004} &  \textbf{0.470$\pm$0.005} \\
            
            & 1-HL & \uline{0.980$\pm$0.000} & \textbf{0.982$\pm$0.000} & 0.969$\pm$0.000 & \uline{0.980$\pm$0.000} & \textbf{0.982$\pm$0.000}  & \textbf{0.982$\pm$0.000} & \textbf{0.982$\pm$0.000} &  \textbf{0.982$\pm$0.000} \\
            
            IAPRTC12 & 1-RL & 0.871$\pm$0.002 & 0.905$\pm$0.001 & 0.842$\pm$0.002 & 0.854$\pm$0.001 & 0.912$\pm$0.003 & 0.918$\pm$0.002 & \uline{0.930$\pm$0.002} & \textbf{0.934$\pm$0.002}\\
            
            & AUC &   0.877$\pm$0.002 &  0.906$\pm$0.001 &  0.843$\pm$0.002 &  0.855$\pm$0.001 &  0.911$\pm$0.003  & 0.918$\pm$0.002 & \uline{0.931$\pm$0.002} &  \textbf{0.935$\pm$0.002} \\
            
            \midrule
            & AP &  0.643$\pm$0.003 & 0.665$\pm$0.003 &  0.493$\pm$0.022 & 0.555$\pm$0.002 & 0.659$\pm$0.004  & 0.681$\pm$0.004 & \uline{0.701$\pm$0.003} &  \textbf{0.712$\pm$0.004} \\
            
            & 1-HL & 0.874$\pm$0.001 & 0.896$\pm$0.001 &  0.840$\pm$0.004 & 0.882$\pm$0.001 & 0.902$\pm$0.001  & 0.906$\pm$0.001 & \uline{0.910$\pm$0.001} &  \textbf{0.911$\pm$0.001} \\
            
            MIRFLICKR & 1-RL & 0.876$\pm$0.006 & 0.894$\pm$0.001 &  0.806$\pm$0.015 & 0.847$\pm$0.001 & 0.895$\pm$0.001  & 0.907$\pm$0.002 & \uline{0.916$\pm$0.001} & \textbf{0.920$\pm$0.001}\\
            
            & AUC &  0.869$\pm$0.005 &  0.880$\pm$0.001 &   0.789$\pm$0.022 &  0.839$\pm$0.001 & 0.876$\pm$0.002 &  0.888$\pm$0.002 & \uline{0.898$\pm$0.001} &  \textbf{0.901$\pm$0.001} \\
            \bottomrule
        \end{tabular}
    }
\end{table*}

\subsection{Hyper-parameter Analysis}
\begin{figure}[!t]
	\centering
        \subfloat[Corel5k]
    {\includegraphics[width=1.9in,height=1.45in]{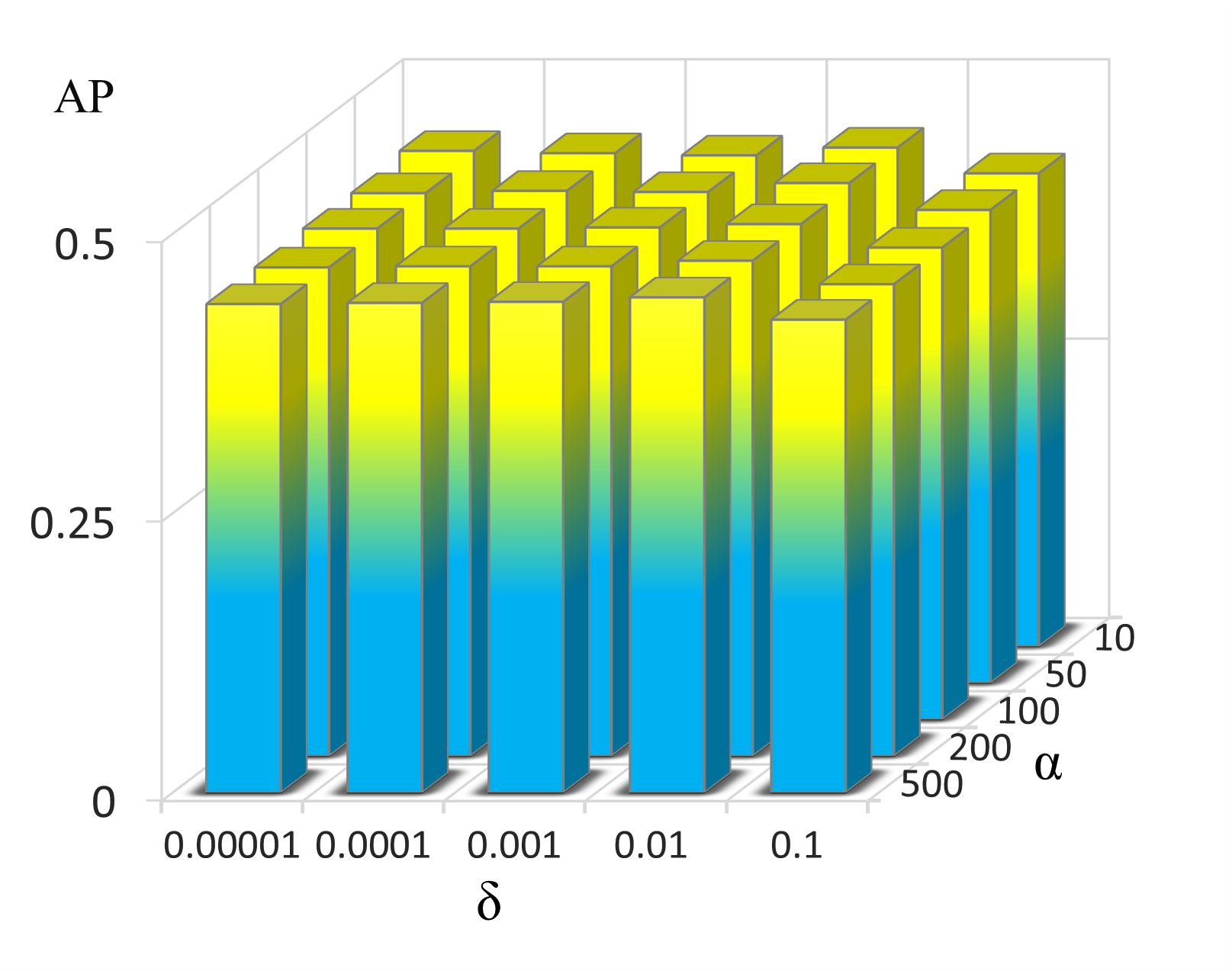}}
	\hfil
	\subfloat[Pascal07]{\includegraphics[width=1.9in,height=1.45in]{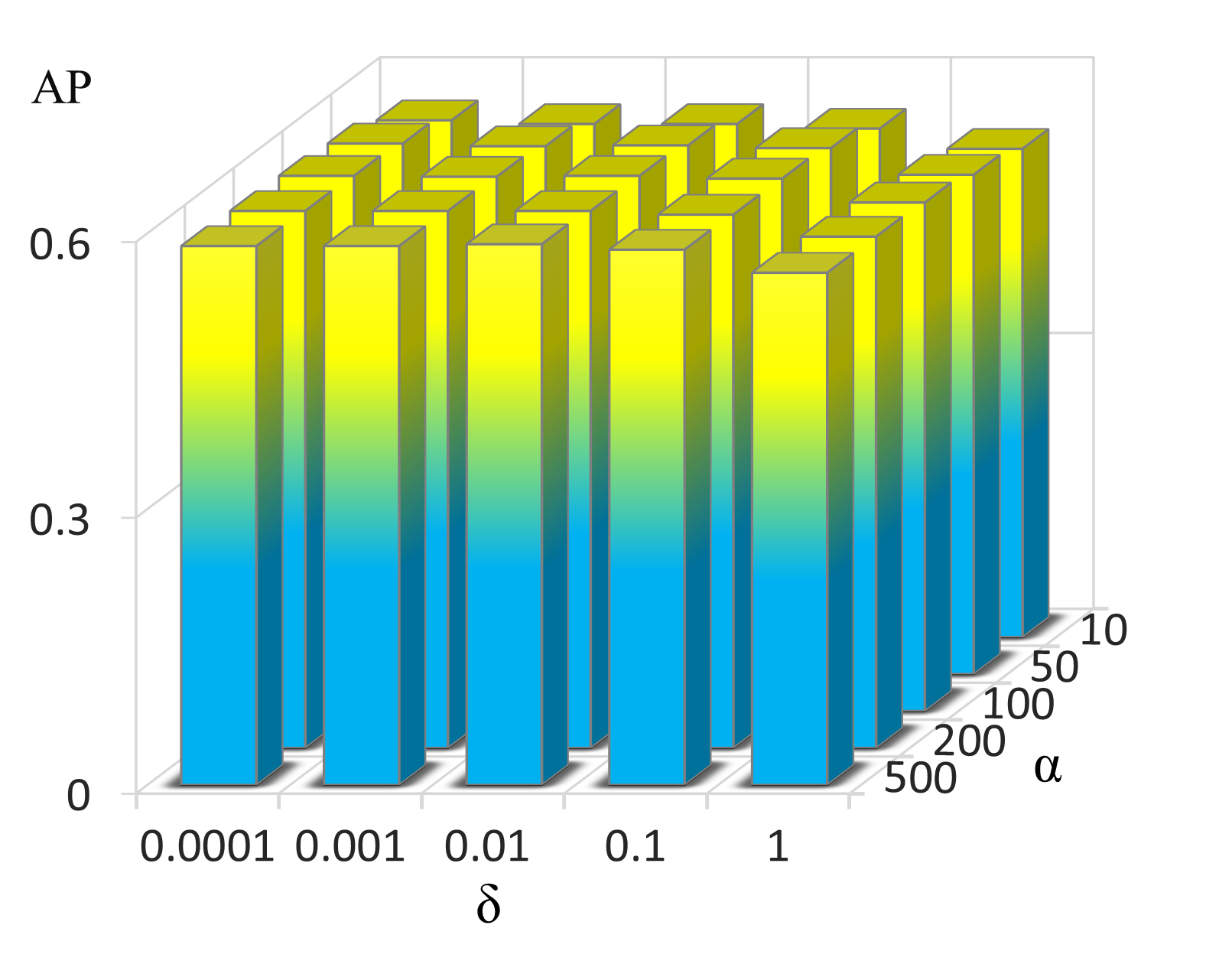}}
        \hfil
	\caption{The hyper-parameter sensitivity experiments on (a) Corkl5k dataset and (b) Pascal07 dataset with 50\% missing instances, 50\% missing labels, and 70\% training samples.}
	\label{figure_five}
\end{figure}

In TACVI-Net, there are two hyper-parameters to be set before training, $\delta $ and $\alpha$. To investigate the model's sensitivity to these two hyperparameters, their values are adjusted within their respective ranges and tested based on the AP metric on both Corel5k and Pascal07 datasets with 50\% missing views, 50\% missing labels, and a 70\% training sample proportion. Throughout the experiments, we strictly controlled the stabilization of other irrelevant hyperparameters to ensure the accuracy of the obtained conclusions and the rigor of the experiments. As shown in Fig.\ref{figure_five}, when $\delta$ is set within the range of [$10^{-5}$, $10^{-2}$] and $\alpha$ varies within [10, 50] for the Corel5k dataset, and for the Pascal07 dataset, where $\delta$ is adjusted to the interval [$10^{-4}$, $10^{-2}$] while $\alpha$ remains within [10, 500], our model remains relatively stable and satisfactory efficiency. By analyzing Figures 5(a) and 5(b) in detail, it can be observed that the optimal hyperparameter combinations for the Corel5k occur when $\delta $ is set to $10^{-2}$ and $\alpha$ to 50, while for the Pascal07, the most optimal configurations are those where $\delta $ takes $10^{-2}$ and $\alpha$ takes 500.

\subsection{Ablation Experiments}
In the ablation study, the performance of each TACVI-Net module is evaluated on three representative datasets. The experiment is conducted under the conditions where 50\% of the instances from each view are utilized, with an additional 50\% of these instances having missing labels, and 70\% of the samples for training. To delve deeper into the effectiveness of individual components, a dual-pronged approach is adopted. Initially, two separable loss functions, $L_{ta}^{l}$ and $L_{re}$, are removed from the foundational model to establish a baseline. Subsequently, each of these losses is reintroduced incrementally to the baseline, allowing for a meticulous examination of their individual contributions to enhancing model performance. In parallel, to comprehensively gauge the efficacy of data imputation strategies, a contrasting experiment is carried out where no data imputation was applied. This direct comparison serve to highlight the profound impact that imputation techniques have on the overall model performance. The relevant experimental results are summarized in Table \ref{table_five}. We can observe that each component of TACVI-Net is vital and beneficial, highlighting the importance of every component in boosting the final classification performance. Furthermore, upon the removal of imputation, there is a noticeable decline in the model’s metrics. This suggests that the imputation strategy indeed addresses issues associated with insufficient information.

\begin{table}
	\centering
	\caption{Ablation experimental results of TACVI-Net on the Corel5k, Pascal07 and ESPGame datasets with 50\% missing views, 50\% missing labels, and 70\% training samples.}
	\label{table_five}
	\setlength{\tabcolsep}{2pt} 
	\scalebox{0.9}{
		\begin{tabular}{cccccccc}
			\toprule
			\multirow{2}{*}{Backbone} & \multirow{2}{*}{$L_{ta}^{l}$} &\multirow{2}{*}{$L_{re}$} &   &Corel5k& Pascal07 & ESPGame\\
			& &  & &  AP  AUC & AP  AUC & AP  AUC\\
			\midrule  
			\checkmark &   & & & 0.338  0.863&0.533 0.839&0.276 0.827\\
		      &   & & & &\\
			\checkmark & \checkmark   & & & 0.375 0.881&0.545 0.845&0.295 0.839 \\
			 &   & & & &\\
			\checkmark &  & \checkmark & &  0.419 0.915&0.571 0.863&0.315 0.858 \\
			 &   & & & &\\
			\checkmark &\checkmark & \checkmark &  &0.447 0.923 &0.588 0.871&0.334 0.866 \\
			\midrule  
			&without imputation& 
			& &0.392 0.907&0.560 0.855 & 0.305 0.848  \\
			\bottomrule
		\end{tabular}
	}
\end{table}

\section{Conclusion}
In this paper, we innovatively introduce a task-augmented cross-view imputation network (TACVI-Net), which departs from previous approaches that rely on assumed missing information to bypass incomplete views. Departing from such conventions, TACVI-Net employs a novel dual-stage strategy which limits the accumulation of redundant information while filling in the missing data so that the model has enough and useful information to learn inter-view consistency and complementarity to improve classification performance. Extensive experiments on five datasets confirm TACVI-Net's superior performance and highlight its innovative capabilities in addressing the challenges of partial multi-view incomplete multi-label classification task.
However, our method does not fully consider the differences in quality among different views. Therefore, future research will more deeply explore how to effectively assess and utilize the quality of each view during the fusion process, to further enhance the robustness and overall performance of multi-view learning.

\section{Acknowledgements}
This work is supported by Guizhou University Basic Research Program ([2024] 04) and Guizhou Provincial Basic Research Program (Natural Science) (No. QianKeHe Basic - [2024] Youth 093).
	 
\bibliographystyle{elsarticle-num-names}
\bibliography{references}

\end{document}